\documentclass{article}

 \usepackage[dblblindworkshop, final]{neurips_2026}

\usepackage[utf8]{inputenc} 
\usepackage[T1]{fontenc}    
\usepackage{hyperref}       
\usepackage{url}            
\usepackage{booktabs}       
\usepackage{amsfonts}       
\usepackage{nicefrac}       
\usepackage{microtype}      
\usepackage{xcolor}         
\usepackage{graphicx}
\usepackage{amsmath}
\title{Before We Adapt: The Zero-Shot Generalization Bottleneck in Continual World Models}
\workshoptitle{Continual World Models}

\author{%
  Amir Mesbah \\
  Independent Researcher \\
  \texttt{mesbahamirhossein@gmail.com}
  \And
  Di Wu$^{1,2}$ \\
  $^{1}$Department of Mechanical, \\ Industrial and Aerospace Engineering \\
  Concordia University \\      
  $^{2}$Department of Electrical and Computer Engineering\\
        McGill University \\
  \texttt{ di.wu@concordia.ca }
}

\begin{document}

\maketitle

\begin{abstract}
Continual reinforcement learning (CRL) requires agents to operate under
environmental changes while reusing prior knowledge whenever possible. We argue
that model-based reinforcement learning provides a promising framework for this
setting because learned world models (WMs) can, in principle, support change
detection, knowledge reuse, adaptation, and acquisition of new knowledge.
However, such benefits depend heavily on whether the learned model can
generalize beyond the contexts in which it was trained. In this work, we propose to formulate this problem with a contextual Markov decision process in which
context is decomposed into behavioral and physical components. This view
organizes continual WM learning into reuse, adaptation, and acquisition
of new knowledge. As a first step, we study direct reuse when behavioral rules
remain unchanged. Using Dyna-Q and TD-MPC in controlled FourRooms tasks, we find that zero-shot reuse can degrade substantially under physical-context changes and that this behavior depends strongly on the observation representation used by
the WM. Additional ablation studies and latent-space diagnostics suggest that
learned representations do not automatically provide the invariances needed for
reuse. These results identify representation generalization as an important
bottleneck for reliable world-model reuse in CRL.
\end{abstract}

\section{Introduction}
Continual reinforcement learning (CRL) considers a setting in which an agent
continuously interacts with its environment and encounters changing tasks or
conditions over time~\citep{khetarpal2022towards,NEURIPS2023_9d8cf124}.
Effective continual agents should not simply relearn after every such change.
Instead, they should preserve previously acquired knowledge and determine
whether it can be reused directly, adapted with limited additional learning,
or whether genuinely new knowledge must be acquired. Achieving this requires
addressing several challenges, including catastrophic forgetting, the
stability-plasticity dilemma, limited interaction budgets, and detecting when
the environment has changed~\citep{ahn2025prevalence, pmlr-v267-tang25g, NEURIPS2025_33c069ac, zuffer2025advancements}.

Model-based reinforcement learning (MBRL) provides a promising foundation for enabling these capabilities through learned world models. By learning an internal world model (WM) of the environment, an agent
can reason about the consequences of its actions and generate imagined
experience, reducing its dependence on costly interaction
~\citep{ha2018world,m2023model}. More importantly, a learned WM can
also provide signals of environmental change when new observations become
inconsistent with its predictions~\citep{zollicoffer2026world}. Furthermore, previously learned dynamics provide a form of knowledge that
should not need to be relearned whenever the environment changes. Recent work has begun to explore world models for CRL, for
example by using online world models ~\citep{pmlr-v267-liu25p}, or considering a generative model as a replay buffer \citep{drago} to reduce catastrophic forgetting across
sequential tasks. However, model-based continual reinforcement
learning remains comparatively less studied than its model-free counterpart
~\citep{zuffer2025advancements}.

To study these capabilities, we need a formulation that captures the
relationships among tasks encountered over time. We view continual learning as
interaction with a \emph{contextual world}, rather than as a sequence of
completely independent Markov decision processes (MDPs). Different tasks
correspond to different contexts and may share different amounts of structure.
A contextual MDP provides a natural way to describe these relationships
~\citep{hallak2015contextual}. This view allows us to distinguish changes in
the underlying behavioral rules, such as action effects, motion rules, or
reward mechanisms, from changes in their physical realization, such as layout,
geometry, or spatial location. How much of this structure remains shared with
previous experience can then guide whether existing knowledge should be reused,
adapted, or replaced by newly learned knowledge.

This perspective makes generalization a prerequisite for knowledge reuse. If a
new task preserves behavioral rules that have already been learned, a world
model should remain useful even when their physical realization changes.
However, this requires its predictions and representations to generalize beyond
the exact states and contexts encountered during training. If they instead
encode context-specific information that is not relevant to the shared
behavioral rules, even a task that should allow direct reuse may require
substantial relearning. Generalization is therefore central to whether
previously learned WM knowledge can actually be reused.

In this work, we develop this perspective using a contextual MDP formulation
that separates behavioral and physical sources of task variation and organizes
continual WM learning into three regimes: \emph{reuse},
\emph{adaptation}, and \emph{acquisition of new knowledge}. We focus
experimentally on the first and most favorable regime, where behavioral rules
remain shared and previously learned dynamics should be reusable without
further learning. Using Dyna-Q~\citep{dyna} and TD-MPC~\citep{hansen2022tdmpc}
in controlled FourRooms tasks, we find that zero-shot reuse can degrade
substantially under physical-context changes and that this behavior depends
strongly on the observation or learned representation. These results identify
the ability to learn context-robust representations as an important requirement
for reliable WM reuse.

\section{Contextual formulation of continual WMs in reinforcement learning}
We model continual reinforcement learning as interaction with a contextual family of environments rather than as a sequence of unrelated MDPs. Following the contextual MDP (CMDP) perspective of \citet{Kirk_2023}, building on \citet{hallak2015contextual}, we consider $\mathcal{M}=(\mathcal{S}',\mathcal{A}',\mathcal{O}',\mathcal{C},\mathcal{P},\mathcal{R},\phi,p_{\mathcal C})$, where $\mathcal{S}'$, $\mathcal{A}'$, and $\mathcal{O}'$ are global state, action, and observation spaces, $\mathcal{C}$ is the context space, $\mathcal{P}$, $\mathcal{R}$, and $\phi$ define the transition, reward, and observation mechanisms across contexts, and $p_{\mathcal C}$ is a distribution over contexts. A context $c_i\in\mathcal{C}$ induces a task $\tau_i=\mathcal{M}(c_i)$, whose task-specific mechanisms are obtained by fixing the context, i.e., $\mathcal{P}i=\mathcal{P}|{c_i}$, $\mathcal{R}i=\mathcal{R}|{c_i}$, and $\phi_i=\phi|_{c_i}$. A continual learning problem then presents a context sequence $\mathbf{c}=(c_1,c_2,\ldots)$ and the corresponding task sequence $\boldsymbol{\tau}=(\tau_1,\tau_2,\ldots)$. We make no independence or identical-distribution assumption on this sequence. The complete formulation is given in Appendix~\ref{app:contextual_formulation}.

We decompose context as $c=(c_B,c_P)\in\mathcal{C}_B\times\mathcal{C}_P$, where $c_B$ denotes \emph{behavioral context} and $c_P$ denotes \emph{physical context}. Behavioral context represents the rules governing how the environment responds to the agent, whereas physical context represents how those rules are instantiated in a particular task, including its geometry, layout, and spatial configuration. This decomposition describes the type of context change, while allowing either context component to affect the transition, reward, or observation mechanisms.

This view suggests three learning regimes for continual world models. When relevant behavioral structure remains valid across a context change, previously learned knowledge may be directly \emph{reused}. When only part of the relevant structure remains applicable, prior knowledge should support efficient \emph{adaptation}. When a context introduces previously unseen behavioral structure, the model must \emph{acquire} new knowledge while preserving previously learned structure. These regimes depend on the structure shared between the current context and prior experience.

In our preliminary study, we examine the reuse regime in a contextual FourRooms environment by changing physical context while keeping behavioral context fixed. A detailed example of this construction is provided in Appendix~\ref{app:fourrooms_example}. We then ask whether learned world models generalize sufficiently across these physical-context changes to support knowledge reuse.

\section{Preliminary study of world models' zero-shot generalization}
\label{sec:generalization}
Zero-shot generalization of world models remains relatively underexplored in
continual model-based RL. Closest to our setting,
\citet{prasanna2024dreamingworldslearningcontextual,roder2026dynamics} study
DreamerV3~\citep{hafner2025dreamerv3} and propose context-aware extensions in
which context is either provided or inferred. Our goal is different. Rather
than focusing on a single architecture or proposing a remedy, we investigate
which factors contribute to poor reuse through ablations over physical context
and observation representation. Our decomposition of physical and behavioral
context also lets us study cases where behavioral rules remain shared while
physical context changes.

\begin{figure}[ht]
    \centering
    \includegraphics[width=0.70\linewidth]{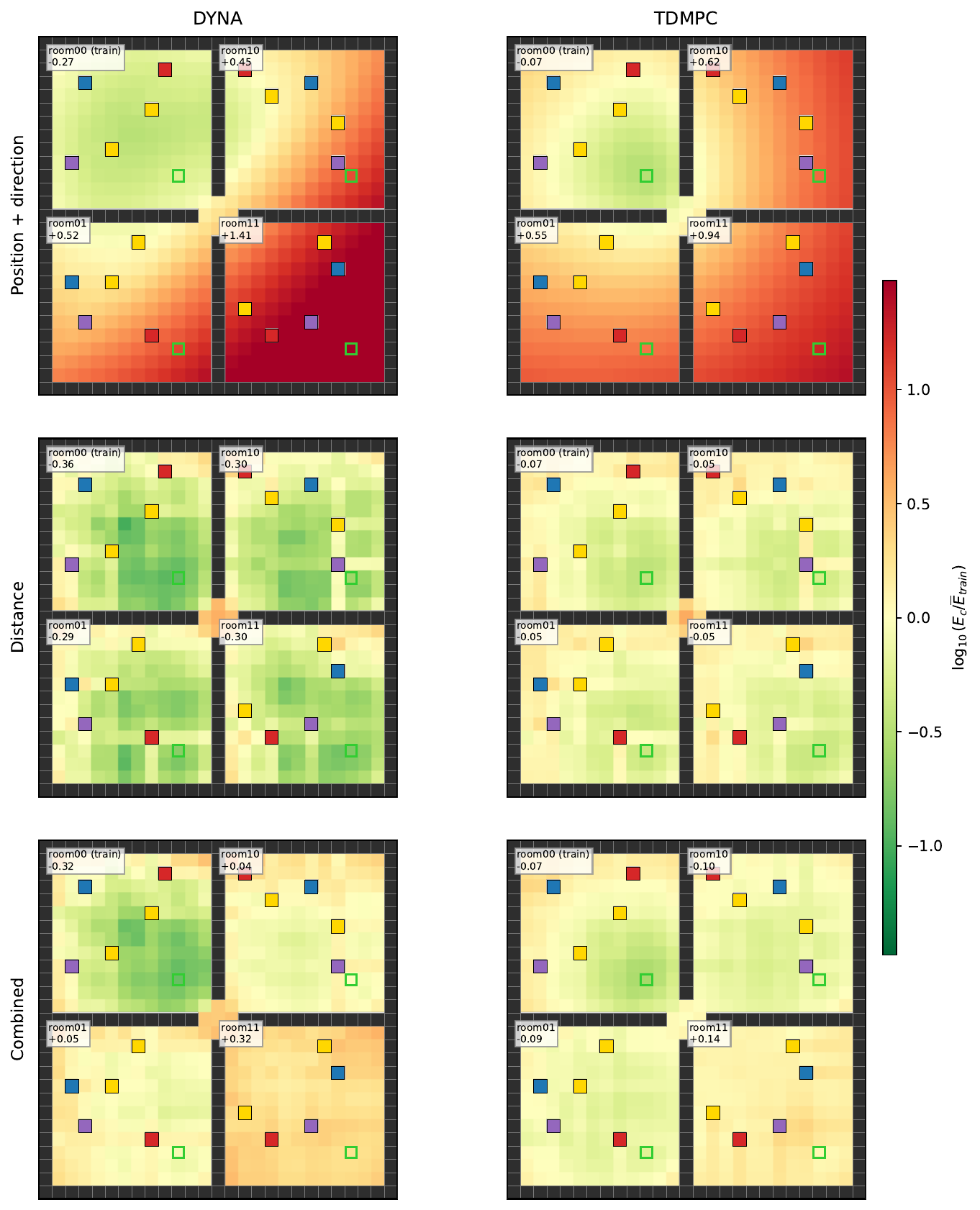}
    \caption{
    Zero-shot one-step dynamics generalization across the four FourRooms
contexts for three observation representations. Results are shown for
Dyna-Q and TD-MPC and averaged over actions, agent directions, and six
seeds. Colors indicate the log generalization gap relative to each model's
training-context error. \emph{Position + direction} contains the agent's
global position and orientation, \emph{distance} contains local distances
from obstacles in the agent's relative directions, and \emph{combined}
contains both types of information.}
    \label{fig:representation_ablation}
\end{figure}

We evaluate Dyna-Q~\citep{dyna} and TD-MPC~\citep{hansen2022tdmpc}, training
each model in one FourRooms context and evaluating it zero-shot in the other
rooms without parameter updates. We report the log generalization gap
$G_c=\log_{10}(E_c/\overline{E}_{\mathrm{train}})$, where prediction error is
normalized by the corresponding mean training-context error; positive values
indicate degradation relative to the training baseline. We compare three
observation representations to examine how representation affects reuse.
Full environment, training, and evaluation details are provided in
Appendix~\ref{app:experimental_details}.

Figure~\ref{fig:representation_ablation} shows that zero-shot generalization
depends strongly on how the environment is represented. With \emph{position +
direction}, both methods show increased prediction error in unseen rooms,
especially as observations move further outside the training range. Part of this degradation may arise from the representation itself, since
spatial translation changes the global coordinates even when the local
dynamics remain unchanged. In contrast,
distance-based observations remain substantially more stable, while the combined
representation shows intermediate behavior. This suggests that sensitivity to
physical-context variation is an important contributor to the reuse bottleneck.
The distance representation should, however, be viewed as an optimistic
reference because it is designed to be less sensitive to the spatial
translations considered here. Real observations are more likely to mix
context-stable and context-dependent information, which is approximated by the
combined representation.

The FourRooms comparison changes both spatial location and layout, so it does
not fully separate coordinate extrapolation from changes in local geometry.
Appendix~\ref{ablation} therefore includes controlled translation-only and
layout-only changes. The translation conditions provide a cleaner test of reuse
by keeping the layout and behavioral rules fixed. Some error under layout
changes is expected because new obstacle locations are not explicitly available
from position-based observations. From a continual learning perspective, this
is still the most favorable regime, since the relevant behavioral knowledge has
already been learned. Our results therefore suggest that successful reuse
depends strongly on the representation. This issue also appears in learned
representations. TD-MPC diagnostics in Appendix~\ref{linear} show that linear probes trained on
the training context do not always transfer to new contexts, and that
representations of matched states can become less aligned after a context shift. Reliable world model
reuse therefore requires representations that preserve shared behavioral
structure while being less sensitive to irrelevant physical variation.

\section{Discussion and conclusion}
We view continual world model learning through three regimes,
\emph{reuse}, \emph{adaptation}, and \emph{acquisition of new knowledge}.
Our contextual MDP view provides a common way to describe these cases. A task
change can be treated as a change in context, and the amount of useful
structure shared with previous experience can guide how prior knowledge should
be used. In this work, we focused on the most favorable case, where behavioral
rules remain shared, and previously learned dynamics should be reusable. Our
results show that even this case can be difficult when the representation is
sensitive to changes in physical context. More invariant observations
generalize considerably better, while the representations learned by TD-MPC do
not automatically remove this dependence. This suggests that reliable reuse
requires representations that preserve shared behavioral structure while being
less sensitive to physical changes that are not relevant to that structure.

This observation also changes how we think about the role of representation
learning in continual world models. Learning a latent state does not by itself
guarantee that information useful across contexts will be represented in a
stable way. A useful continual representation should preserve the parts of the
environment that remain valid across tasks while allowing context-specific
information to change. Our results therefore motivate further study of
context-invariant or factorized representations, although how such structure
should be learned without removing information needed for prediction remains
an open question. World model prediction errors may also provide useful signals
for detecting when the current context has changed and when previously learned
knowledge is no longer sufficient.

Several important parts of the framework remain to be studied. We plan to
extend the evaluation to DreamerV3 and other modern world model approaches and
to include changes in behavioral context, where zero-shot reuse may no longer
be sufficient. We also plan to study how an agent can detect context changes,
estimate how much previous knowledge is still useful, and adapt when only part
of the learned structure remains valid. When genuinely new knowledge must be
learned, plasticity becomes important, but it must be balanced with preserving
previously acquired knowledge and avoiding negative transfer. Another direction
is to identify which parts of current world model agents, including encoders,
latent dynamics models, recurrent components, and planners, are most
responsible for these limitations. Together, these questions move toward world
models that can make better use of previous experience as their context changes
over time.

\begin{ack}


We thank Benoit Boulet for his valuable support during this work. The authors received no external funding for this work.

\end{ack}
\bibliographystyle{plainnat}
\bibliography{ref}

@article{hansen2022tdmpc,
  title={Temporal Difference Learning for Model Predictive Control},
  author={Hansen, Nicklas and Wang, Xiaolong and Su, Hao},
  journal={International Conference on Machine Learning},
  year={2022}
}

@article{khetarpal2022towards,
  title={Towards continual reinforcement learning: A review and perspectives},
  author={Khetarpal, Khimya and Riemer, Matthew and Rish, Irina and Precup, Doina},
  journal={Journal of Artificial Intelligence Research},
  volume={75},
  pages={1401--1476},
  year={2022}
}

@inproceedings{NEURIPS2023_9d8cf124,
 author = {Abel, David and Barreto, Andre and Van Roy, Benjamin and Precup, Doina and van Hasselt, Hado and Singh, Satinder},
 booktitle = {Advances in Neural Information Processing Systems},
 doi = {10.52202/075280-2192},
 editor = {A. Oh and T. Naumann and A. Globerson and K. Saenko and M. Hardt and S. Levine},
 pages = {50377--50407},
 publisher = {Curran Associates, Inc.},
 title = {A Definition of Continual Reinforcement Learning},
 volume = {36},
 year = {2023}
}

@article{zuffer2025advancements,
  title={Advancements and challenges in continual reinforcement learning: A comprehensive review},
  author={Zuffer, Amara and Burke, Michael and Harandi, Mehrtash},
  journal={arXiv preprint arXiv:2506.21899},
  year={2025}
}

@article{m2023model,
  title={Model-based reinforcement learning: A survey},
  author={M. Moerland, Thomas and Broekens, Joost and Plaat, Aske and M. Jonker, Catholijn},
  journal={Foundations and Trends in Machine Learning},
  volume={16},
  number={1},
  pages={1--118},
  year={2023},
  publisher={Emerald Publishing Limited}
}

@article{ha2018world,
  title={World models},
  author={Ha, David and Schmidhuber, J{\"u}rgen},
  journal={arXiv preprint arXiv:1803.10122},
  volume={2},
  number={3},
  pages={440},
  year={2018}
}

@inproceedings{zollicoffer2026world,
  title={World model robustness via surprise recognition},
  author={Zollicoffer, Geigh and Chopra, Tanush and Yan, Mingkuan and Ma, Xiaoxu and Eaton, Kenneth and Riedl, Mark},
  booktitle={Proceedings of the IEEE/CVF Conference on Computer Vision and Pattern Recognition},
  pages={3146--3155},
  year={2026}
}

@article{hallak2015contextual,
  title={Contextual markov decision processes},
  author={Hallak, Assaf and Di Castro, Dotan and Mannor, Shie},
  journal={arXiv preprint arXiv:1502.02259},
  year={2015}
}

@article{hafner2025dreamerv3,
  title={Mastering diverse control tasks through world models},
  author={Hafner, Danijar and Pasukonis, Jurgis and Ba, Jimmy and Lillicrap, Timothy},
  journal={Nature},
  pages={1--7},
  year={2025},
  publisher={Nature Publishing Group}
}

@article{dyna,
author = {Sutton, Richard S.},
title = {Dyna, an integrated architecture for learning, planning, and reacting},
year = {1991},
issue_date = {Aug. 1991},
publisher = {Association for Computing Machinery},
address = {New York, NY, USA},
volume = {2},
number = {4},
issn = {0163-5719},
doi = {10.1145/122344.122377},
journal = {SIGART Bull.},
month = jul,
pages = {160–163},
numpages = {4}
}

@misc{leworldmodel2026,
      title={LeWorldModel: Stable End-to-End Joint-Embedding Predictive Architecture from Pixels}, 
      author={Lucas Maes and Quentin Le Lidec and Damien Scieur and Yann LeCun and Randall Balestriero},
      year={2026},
      eprint={2603.19312},
      archivePrefix={arXiv},
      primaryClass={cs.LG},
      url={https://arxiv.org/abs/2603.19312}, 
}

@misc{janner2021trustmodelmodelbasedpolicy,
      title={When to Trust Your Model: Model-Based Policy Optimization}, 
      author={Michael Janner and Justin Fu and Marvin Zhang and Sergey Levine},
      year={2021},
      eprint={1906.08253},
      archivePrefix={arXiv},
      primaryClass={cs.LG},
       
}

@misc{prasanna2024dreamingworldslearningcontextual,
      title={Dreaming of Many Worlds: Learning Contextual World Models Aids Zero-Shot Generalization}, 
      author={Sai Prasanna and Karim Farid and Raghu Rajan and André Biedenkapp},
      year={2024},
      eprint={2403.10967},
      archivePrefix={arXiv},
      primaryClass={cs.LG}, 
}

@misc{lee2020contextawaredynamicsmodelgeneralization,
      title={Context-aware Dynamics Model for Generalization in Model-Based Reinforcement Learning}, 
      author={Kimin Lee and Younggyo Seo and Seunghyun Lee and Honglak Lee and Jinwoo Shin},
      year={2020},
      eprint={2005.06800},
      archivePrefix={arXiv},
      primaryClass={cs.LG},
     
}

@misc{shimizu2024bisimulationmetricmodelpredictive,
      title={Bisimulation metric for Model Predictive Control}, 
      author={Yutaka Shimizu and Masayoshi Tomizuka},
      year={2024},
      eprint={2410.04553},
      archivePrefix={arXiv},
      primaryClass={cs.LG},
      
}

@misc{beukman2023dynamicsgeneralisationreinforcementlearning,
      title={Dynamics Generalisation in Reinforcement Learning via Adaptive Context-Aware Policies}, 
      author={Michael Beukman and Devon Jarvis and Richard Klein and Steven James and Benjamin Rosman},
      year={2023},
      eprint={2310.16686},
      archivePrefix={arXiv},
      primaryClass={cs.AI},
     
}

@article{Kirk_2023,
   title={A Survey of Zero-shot Generalisation in Deep Reinforcement Learning},
   volume={76},
   ISSN={1076-9757},
   
   DOI={10.1613/jair.1.14174},
   journal={Journal of Artificial Intelligence Research},
   publisher={AI Access Foundation},
   author={Kirk, Robert and Zhang, Amy and Grefenstette, Edward and Rocktäschel, Tim},
   year={2023},
   month=Jan, pages={201–264} }

@inproceedings{huang2021continual,
  title={Continual Model-Based Reinforcement Learning with Hypernetworks},
  author={Huang, Yizhou and Xie, Kevin and Bharadhwaj, Homanga and Shkurti, Florian},
  booktitle={IEEE International Conference on Robotics and Automation (ICRA)},
  year={2021}
}

@article{fu2022model,
  title={Model-based lifelong reinforcement learning with bayesian exploration},
  author={Fu, Haotian and Yu, Shangqun and Littman, Michael and Konidaris, George},
  journal={Advances in Neural Information Processing Systems},
  volume={35},
  pages={32369--32382},
  year={2022}
}

@article{roder2026dynamics,
  title={Dynamics-aligned latent imagination in contextual world models for zero-shot generalization},
  author={R{\"o}der, Frank and Benad, Jan and Eppe, Manfred and Banerjee, Pradeep},
  journal={Advances in Neural Information Processing Systems},
  volume={38},
  pages={72736--72771},
  year={2026}
}

@InProceedings{drago,
  title = 	 {Knowledge Retention in Continual Model-Based Reinforcement Learning},
  author =       {Fu, Haotian and Sun, Yixiang and Littman, Michael and Konidaris, George},
  booktitle = 	 {Proceedings of the 42nd International Conference on Machine Learning},
  pages = 	 {17832--17851},
  year = 	 {2025},
  editor = 	 {Singh, Aarti and Fazel, Maryam and Hsu, Daniel and Lacoste-Julien, Simon and Berkenkamp, Felix and Maharaj, Tegan and Wagstaff, Kiri and Zhu, Jerry},
  volume = 	 {267},
  series = 	 {Proceedings of Machine Learning Research},
  month = 	 {13--19 Jul},
  publisher =    {PMLR},
  
}

@article{
alyahya2026arrow,
title={{ARROW}: Augmented Replay for {RO}bust World models},
author={Abdulaziz Alyahya and Abdallah Al Siyabi and Markus R. Ernst and Luke Yang and Levin Kuhlmann and Gideon Kowadlo},
journal={Transactions on Machine Learning Research},
issn={2835-8856},
year={2026},
}

@inproceedings{rahimi2023replay,
  title={Replay buffer with local forgetting for adapting to local environment changes in deep model-based reinforcement learning},
  author={Rahimi-Kalahroudi, Ali and Rajendran, Janarthanan and Momennejad, Ida and van Seijen, Harm and Chandar, Sarath},
  booktitle={Conference on Lifelong Learning Agents},
  pages={21--42},
  year={2023},
  organization={PMLR}
}

@InProceedings{pmlr-v232-kessler23a,
  title = 	 {The Effectiveness of World Models for Continual Reinforcement Learning},
  author = {Kessler, Samuel and Ostaszewski, Mateusz and Bortkiewicz, Micha{\l} Pawe{\l} and \.{Z}arski, Mateusz and Wolczyk, Maciej and Parker-Holder, Jack and Roberts, Stephen J. and Mi{\l}o{\'s}, Piotr},
  booktitle = 	 {Proceedings of The 2nd Conference on Lifelong Learning Agents},
  pages = 	 {184--204},
  year = 	 {2023},
  editor = 	 {Chandar, Sarath and Pascanu, Razvan and Sedghi, Hanie and Precup, Doina},
  volume = 	 {232},
  series = 	 {Proceedings of Machine Learning Research},
  month = 	 {22--25 Aug},
  publisher =    {PMLR},
  }

@inproceedings{schwarz2018progress,
  title={Progress \& compress: A scalable framework for continual learning},
  author={Schwarz, Jonathan and Czarnecki, Wojciech and Luketina, Jelena and Grabska-Barwinska, Agnieszka and Teh, Yee Whye and Pascanu, Razvan and Hadsell, Raia},
  booktitle={International conference on machine learning},
  pages={4528--4537},
  year={2018},
  organization={PMLR}
}

@article{kirkpatrick2017overcoming,
  title={Overcoming catastrophic forgetting in neural networks},
  author={Kirkpatrick, James and Pascanu, Razvan and Rabinowitz, Neil and Veness, Joel and Desjardins, Guillaume and Rusu, Andrei A and Milan, Kieran and Quan, John and Ramalho, Tiago and Grabska-Barwinska, Agnieszka and others},
  journal={Proceedings of the national academy of sciences},
  volume={114},
  number={13},
  pages={3521--3526},
  year={2017},
  publisher={National Academy of Sciences}
}

@InProceedings{pmlr-v267-tang25g,
  title = 	 {Mitigating Plasticity Loss in Continual Reinforcement Learning by Reducing Churn},
  author =       {Tang, Hongyao and Obando-Ceron, Johan and Castro, Pablo Samuel and Courville, Aaron and Berseth, Glen},
  booktitle = 	 {Proceedings of the 42nd International Conference on Machine Learning},
  pages = 	 {58883--58904},
  year = 	 {2025},
  editor = 	 {Singh, Aarti and Fazel, Maryam and Hsu, Daniel and Lacoste-Julien, Simon and Berkenkamp, Felix and Maharaj, Tegan and Wagstaff, Kiri and Zhu, Jerry},
  volume = 	 {267},
  series = 	 {Proceedings of Machine Learning Research},
  month = 	 {13--19 Jul},
  publisher =    {PMLR},
  
}

@InProceedings{pmlr-v267-liu25p,
  title = 	 {Continual Reinforcement Learning by Planning with Online World Models},
  author =       {Liu, Zichen and Fu, Guoji and Du, Chao and Lee, Wee Sun and Lin, Min},
  booktitle = 	 {Proceedings of the 42nd International Conference on Machine Learning},
  pages = 	 {38397--38423},
  year = 	 {2025},
  editor = 	 {Singh, Aarti and Fazel, Maryam and Hsu, Daniel and Lacoste-Julien, Simon and Berkenkamp, Felix and Maharaj, Tegan and Wagstaff, Kiri and Zhu, Jerry},
  volume = 	 {267},
  series = 	 {Proceedings of Machine Learning Research},
  month = 	 {13--19 Jul},
  publisher =    {PMLR},
}

@inproceedings{
ahn2025prevalence,
title={Prevalence of Negative Transfer in Continual Reinforcement Learning: Analyses and a Simple Baseline},
author={Hongjoon Ahn and Jinu Hyeon and Youngmin Oh and Bosun Hwang and Taesup Moon},
booktitle={The Thirteenth International Conference on Learning Representations},
year={2025},
}

@inproceedings{NEURIPS2025_33c069ac,
 author = {Hu, Jinwu and Lian, ZiHao and Wen, Zhiquan and Li, Chenghao and Chen, Guohao and Wen, Xutao and Xiao, Bin and Tan, Mingkui},
 booktitle = {Advances in Neural Information Processing Systems},
 doi = {10.52202/085713-1219},
 editor = {D. Belgrave and C. Zhang and H. Lin and R. Pascanu and P. Koniusz and M. Ghassemi and N. Chen},
 pages = {36283--36313},
 publisher = {Curran Associates, Inc.},
 title = {Continual Knowledge Adaptation for Reinforcement Learning},
 volume = {38, Main Conference},
 year = {2025}
}


\appendix

\section{Detailed definition of contextual continual world model formulation}
\label{app:contextual_formulation}
We consider a contextual family of environments $\mathcal{M}=\langle \mathcal{S}',\mathcal{A}',\mathcal{O}',\mathcal{C}, \mathcal{P},\mathcal{R},\phi,p_{\mathcal C} \rangle$, where $\mathcal{S}'$, $\mathcal{A}'$, and $\mathcal{O}'$ are the global state, action, and observation spaces, $\mathcal{C}$ is the context space, $\mathcal{P}$, $\mathcal{R}$, and $\phi$ define the transition, reward, and observation mechanisms across contexts, and $p_{\mathcal C}$ is a distribution over contexts.

We decompose the context as $c=(c_B,c_P)\in\mathcal{C}_B\times\mathcal{C}_P$, where $c_B$ is the \emph{behavioral context} and $c_P$ is the \emph{physical context}. Behavioral context describes the rules governing how the environment responds to the agent, while physical context describes how these rules are instantiated in a particular environment. Either type of context change may affect the transition, reward, or observation mechanisms. 

The state, action, and observation spaces of a task can depend on its context. We represent them using the set-valued mappings $\mathcal{S}:\mathcal{C}\rightarrow 2^{\mathcal{S}'}$, $\mathcal{A}:\mathcal{C}\rightarrow 2^{\mathcal{A}'}$, and $\mathcal{O}:\mathcal{C}\rightarrow 2^{\mathcal{O}'}$. For a context $c_i$, the corresponding task-specific spaces are therefore $\mathcal{S}_i=\mathcal{S}(c_i)\subseteq\mathcal{S}'$, $\mathcal{A}_i=\mathcal{A}(c_i)\subseteq\mathcal{A}'$, and $\mathcal{O}_i=\mathcal{O}(c_i)\subseteq\mathcal{O}'$. These spaces do not need to be disjoint, so different tasks may share states, actions, or observations. A context $c_i$ induces a task $\tau_i=\mathcal{M}(c_i)$, which we write as
$$
\tau_i=
\left\langle
\mathcal{S}_i,\mathcal{A}_i,\mathcal{O}_i,
\mathcal{P}_i,\mathcal{R}_i,\phi_i
\right\rangle .
$$
The task-specific mechanisms are obtained by fixing the context: $\mathcal{P}_i(\cdot\mid s,a)=\mathcal{P}(\cdot\mid s,a,c_i)$, $\mathcal{R}_i(s,a,s')=\mathcal{R}(s,a,s',c_i)$, and $\phi_i(s)=\phi(s,c_i)$. Thus, the context determines both the spaces that are valid for a task and its transition, reward, and observation mechanisms.

A continual learning problem consists of a context sequence $\mathbf{c}=(c_1,c_2,\ldots,c_N)$, with $c_i\in\mathcal{C}$ and $N\in\mathbb{N}\cup\{\infty\}$, which induces the task sequence $\boldsymbol{\tau}=(\tau_1,\tau_2,\ldots,\tau_N)$, where $\tau_i=\mathcal{M}(c_i)$. We do not assume that the contexts are independent or identically distributed. More generally, the context at task $i$ may depend on previous contexts through $p(c_i\mid c_{1:i-1})$. For a finite task sequence, this can be written as \[ p_{\mathrm{seq}}(c_{1:N}) = p(c_1)\prod_{i=2}^{N}p(c_i\mid c_{1:i-1}). \] Independent sampling from the CMDP context distribution is a special case in which $p(c_i\mid c_{1:i-1})=p_{\mathcal C}(c_i)$. 
Since $c_i=(c_{B,i},c_{P,i})$, a change between task sequences can occur in three ways: \[ \begin{aligned} c_{B,i+1}\neq c_{B,i},\quad c_{P,i+1}=c_{P,i} &\qquad \text{(behavioral change)},\\ c_{B,i+1}=c_{B,i},\quad c_{P,i+1}\neq c_{P,i} &\qquad \text{(physical change)},\\ c_{B,i+1}\neq c_{B,i},\quad c_{P,i+1}\neq c_{P,i} &\qquad \text{(joint change)}. \end{aligned} \] The transition, reward, and observation functions may in general depend on both context components, i.e., $\mathcal{P}(\cdot\mid s,a,c_B,c_P)$, $\mathcal{R}(s,a,s',c_B,c_P)$, and $\phi(s,c_B,c_P)$. However, a change in context does not require all of these functions to change. For example, a function can remain unchanged with respect to physical context if its value is the same for different $c_P$ while $c_B$ is fixed, and similarly for behavioral context. Therefore, each task change may affect only some of the task-specific spaces or functions, while the remaining parts can be shared across tasks.

\section{FourRooms contextual-world example}
\label{app:fourrooms_example}

\begin{figure}[ht]
    \centering
    \includegraphics[width=0.55\linewidth]{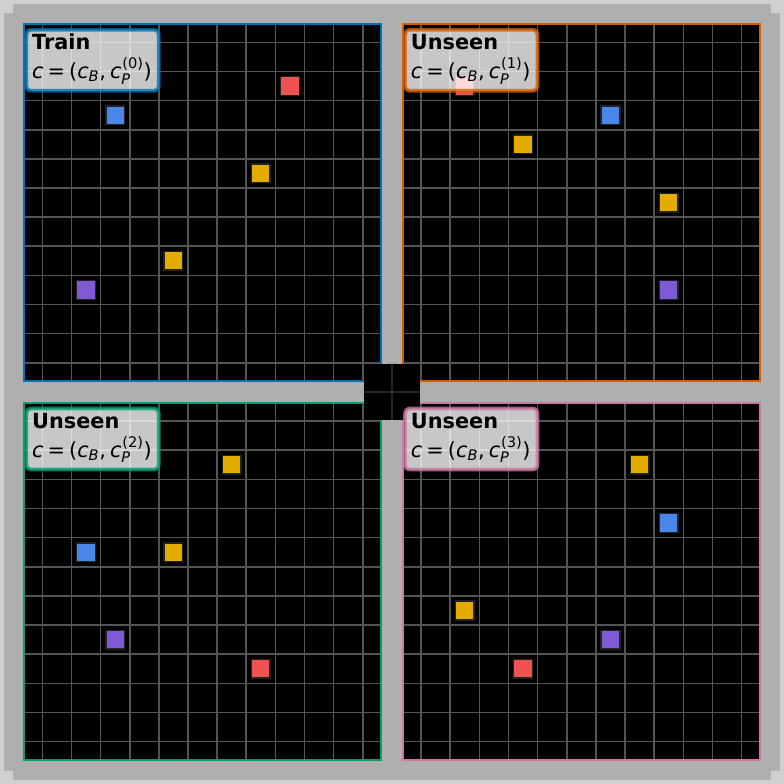}
    \caption{
    Contextual FourRooms example. Tasks may differ in physical context $c_P$
    while sharing behavioral rules $c_B$. When the relevant behavioral
    structure remains applicable, this corresponds to the \emph{reuse} regime in which
    previously learned knowledge should remain useful rather than being learned
    again from scratch.
    }
    \label{fig:context_reuse_appendix}
\end{figure}

Figure~\ref{fig:context_reuse_appendix} illustrates how the contextual-world
formulation is instantiated in FourRooms. Each room can be viewed as a task
with its own physical context $c_P$, while the underlying behavioral context
$c_B$ can remain shared. Physical context may vary through spatial location,
layout, or obstacle configuration, whereas the rules governing actions and
rewards remain unchanged. For example, the action \emph{forward} follows the
same rule in every context. The agent moves one cell when the target cell is
free and remains in place when it is blocked.

\section{Experimental details}
\label{app:experimental_details}
\paragraph{Environment.}
We use a contextual version of the FourRooms environment. The environment and
the \emph{position + direction} observation follow the continual model-based RL
setup of \citet{drago}, while we modify the task contexts and evaluation
procedure to study zero-shot generalization. The transition dynamics are
deterministic. The agent can turn left, turn right, or move forward by one grid
cell; forward movement succeeds when the target cell is free and otherwise
leaves the agent in place. We use the same dense reward mechanism across
contexts,
$r_t=1-\lVert p_t-p_g\rVert_2/d_{\mathrm{goal}}$, where
$d_{\mathrm{goal}}=12$. Thus, action semantics, motion rules, and the reward
mechanism remain shared while physical properties such as spatial location,
layout, and obstacle configuration may vary.

We evaluate three observation representations. The \emph{position + direction}
observation is
$
o_t^{\mathrm{pos}}
=
(\tilde{x}_t,\tilde{y}_t,\cos\theta_t,\sin\theta_t),
$
where $(x_t,y_t)$ is the agent's global position and $\theta_t$ its orientation.
The global coordinates are min--max normalized to $[-1,1]$ using fixed global
bounds $[-26,130]$ and clipped to this range. The \emph{distance} observation is
$
o_t^{\mathrm{dist}}
=
(\cos\theta_t,\sin\theta_t,
d_t^{\mathrm{front}},
d_t^{\mathrm{back}},
d_t^{\mathrm{left}},
d_t^{\mathrm{right}}),
$
where the four distances are measured to the nearest obstacle or room boundary
in each relative direction and normalized by $2r$, with room size $r=13$.
The \emph{combined} observation is
$
o_t^{\mathrm{comb}}
=
(\tilde{x}_t,\tilde{y}_t,\cos\theta_t,\sin\theta_t,
d_t^{\mathrm{front}},
d_t^{\mathrm{back}},
d_t^{\mathrm{left}},
d_t^{\mathrm{right}}),
$
and therefore contains both global-position and local-distance information.
The distance representation is less sensitive to global spatial translation,
while the combined representation mixes both global and local information.
The position + direction representation is not invariant to spatial
translation. Under this observation, translating the environment changes $(x_t,y_t)$ and may move observations outside the range seen during training, even when the underlying behavioral rules remain unchanged. Thus, some degradation under translation
may arise from the representation choice itself. Under position + direction
observations, the locations of newly introduced obstacles are also not
explicitly provided to the model.

\paragraph{Training protocol.}
Dyna-Q and TD-MPC are trained only in the \texttt{room00} context. For each
algorithm and observation representation, we train six models using random
seeds $\{1,2,43,100,256,314\}$ for $100{,}000$ environment steps. During evaluation, all model
parameters are frozen, and no adaptation or fine-tuning is performed in the
test contexts. The hyperparameters used for both methods are summarized in
Table~\ref{tab:hyperparameters}. The planning horizon remains fixed at $3$ throughout training.

\begin{table}[t]
\centering
\small
\caption{Training and model hyperparameters used for Dyna and TD-MPC.}
\label{tab:hyperparameters}
\begin{tabular}{ll}
\toprule
\textbf{Hyperparameter} & \textbf{Value} \\
\midrule

\multicolumn{2}{l}{\textbf{Dyna}} \\
\midrule
Total timesteps                & $100{,}000$ \\
Q-network learning rate        & $3\times10^{-4}$ \\
Dynamics-model learning rate   & $1\times10^{-3}$ \\
Learning starts                & $5{,}000$ \\
Training frequency             & $4$ \\
Batch size                     & $128$ \\
Discount factor ($\gamma$)     & $0.99$ \\
Target-network update frequency & $1{,}000$ \\
Replay-buffer capacity         & $100{,}000$ \\
Initial exploration rate       & $1.0$ \\
Final exploration rate         & $0.05$ \\
Exploration fraction           & $0.2$ \\
Planning steps                 & $3$ \\
Q-network hidden dimensions    & $[120,84]$ \\
Dynamics-model hidden dimensions & $[128,128]$ \\

\midrule
\multicolumn{2}{l}{\textbf{TD-MPC}} \\
\midrule
Total timesteps                   & $100{,}000$ \\
Episode-buffer capacity           & $10{,}000$ \\
Transition-buffer capacity        & $100{,}000$ \\
Latent dimension                  & $50$ \\
MLP hidden dimension              & $256$ \\
Warm-up steps                     & $5{,}000$ \\
Planning horizon                  & $3 (fixed)$ \\
Planning samples                  & $512$ \\
Planning elites                   & $32$ \\
Planning optimization iterations  & $6$ \\
Mixture coefficient               & $0.05$ \\
Temperature                       & $0.5$ \\
Momentum                          & $0.1$ \\
Minimum standard deviation        & $0.05$ \\
Standard-deviation schedule       & $\mathrm{linear}(1.0,0.1,250000)$ \\
Batch size                        & $128$ \\
Learning rate                     & $1\times10^{-4}$ \\
Discount factor                   & $0.99$ \\
Update frequency                  & $1{,}000$ \\
Target update coefficient ($\tau$)& $0.01$ \\
Gradient clipping norm            & $1000$ \\
$\rho$                             & $0.99$ \\
Consistency-loss coefficient      & $1.0$ \\
Reward-loss coefficient           & $1.0$ \\
Value-loss coefficient            & $1.0$ \\

\bottomrule
\end{tabular}
\end{table}

\paragraph{Linear schedules.}
We use the notation $\mathrm{linear}(a,b,T)$ for a schedule that linearly
interpolates from $a$ to $b$ over $T$ environment steps and remains at $b$
afterward. The standard-deviation schedule is
$\mathrm{linear}(1.0,0.1,250000)$. Since training lasts $100000$ environment
steps, only the first part of this schedule is used.

\paragraph{Generalization regimes.}
We organize the four base rooms according to how their spatial observation support relates to the training room. \texttt{room00} is the \emph{in-distribution reference}, since it is the context used during training. \texttt{room10} and \texttt{room01} form an \emph{intermediate} regime where one spatial coordinate lies in a range related to the training context while the other moves outside it. Finally, \texttt{room11} provides the \emph{full extrapolation} case, in which both spatial dimensions lie outside the training range. The layouts also differ between these rooms; controlled translation-only and layout-only experiments in Appendix~\ref{ablation} are used for more investigation.

\paragraph{One-step dynamics evaluation.}
For every valid state in an evaluation context, we test the three actions
$\{\text{left},\text{right},\text{forward}\}$ for all four agent directions.
For Dyna-Q, next-state prediction error is measured in observation space.
TD-MPC predicts in its learned latent space, so its predicted latent next state
is compared with the encoding of the true next observation. We use mean
squared error (MSE) as the one-step dynamics error. Because the two methods
operate in different representation spaces, we do not compare their raw MSE
values directly.

\paragraph{Generalization gap.}
For each trained run and action, we first compute the mean one-step dynamics
error over the complete training context,
$\overline{E}_{\mathrm{train}}$. For an evaluated state with error $E_c$, we
report
$G_c=\log_{10}(E_c/\overline{E}_{\mathrm{train}})$.
Positive values indicate higher error than the training-context average and
negative values indicate lower error. Importantly, individual cells in the
training-context heatmap are not expected to equal zero because each cell is
normalized by the \emph{mean} error over the whole training context rather
than by its own error. For the reported heatmaps, values are averaged over
agent directions and seeds and then equally across the three actions.

\paragraph{Evaluation contexts.}
In addition to the four base FourRooms tasks shown in the main paper, we use
controlled physical-context changes to separate different sources of
generalization error. Positions are expressed in grid-cell coordinates, and
each room has size $r=13$. Context offsets are converted to global-coordinate
offsets by multiplying them by $r$. Thus, adjacent base rooms are separated by
$13$ grid cells along the corresponding axis. The two translation-only
conditions preserve the training layout, obstacles, and behavioral rules while
shifting the global coordinates by $2r=26$ and $3r=39$ grid cells along the
$x$-axis, respectively. The \emph{layout} condition changes the obstacle layout
while keeping the original spatial location. Finally, the \emph{combined}
(layout + translation) condition changes both. The translation-only conditions
therefore isolate sensitivity to global spatial translation while preserving
the local environment structure.

\section{Complementary generalization results}
\label{app:complementary}

\subsection{Controlled physical-context shifts}
\label{ablation}

\begin{figure}[ht]
    \centering
    \includegraphics[width=\linewidth]
{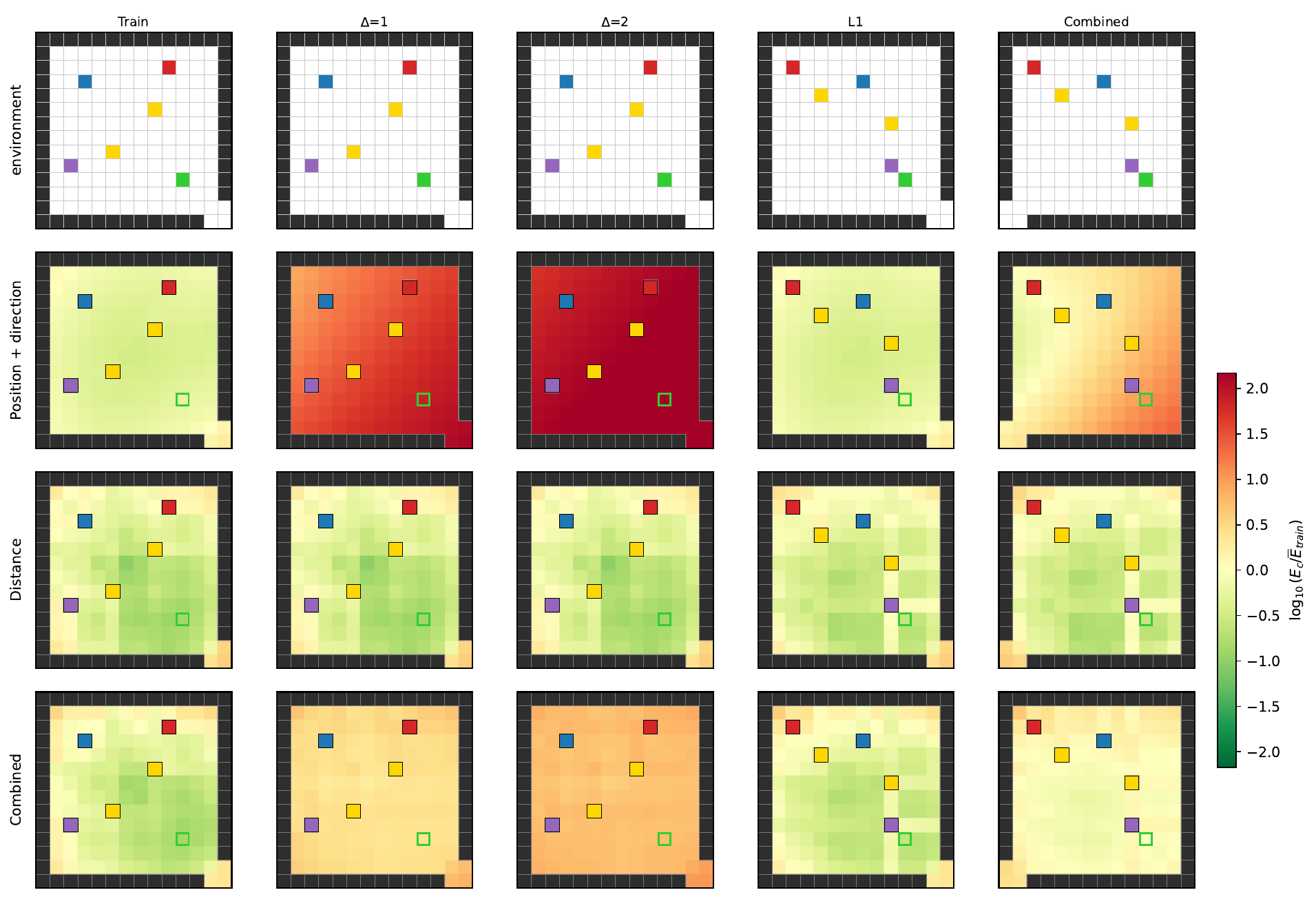}
    \caption{
    Controlled physical-context shifts for Dyna-Q across the three observation
    representations. Translation conditions preserve the training layout while
    changing its global spatial realization; the layout condition changes local
    geometry at the training location; and layout + translation changes both.
    Colors show the one-step dynamics generalization gap.
    }
    \label{fig:appendix_dyna_contexts}
\end{figure}

\begin{figure}[ht]
    \centering
    \includegraphics[width=\linewidth]
    {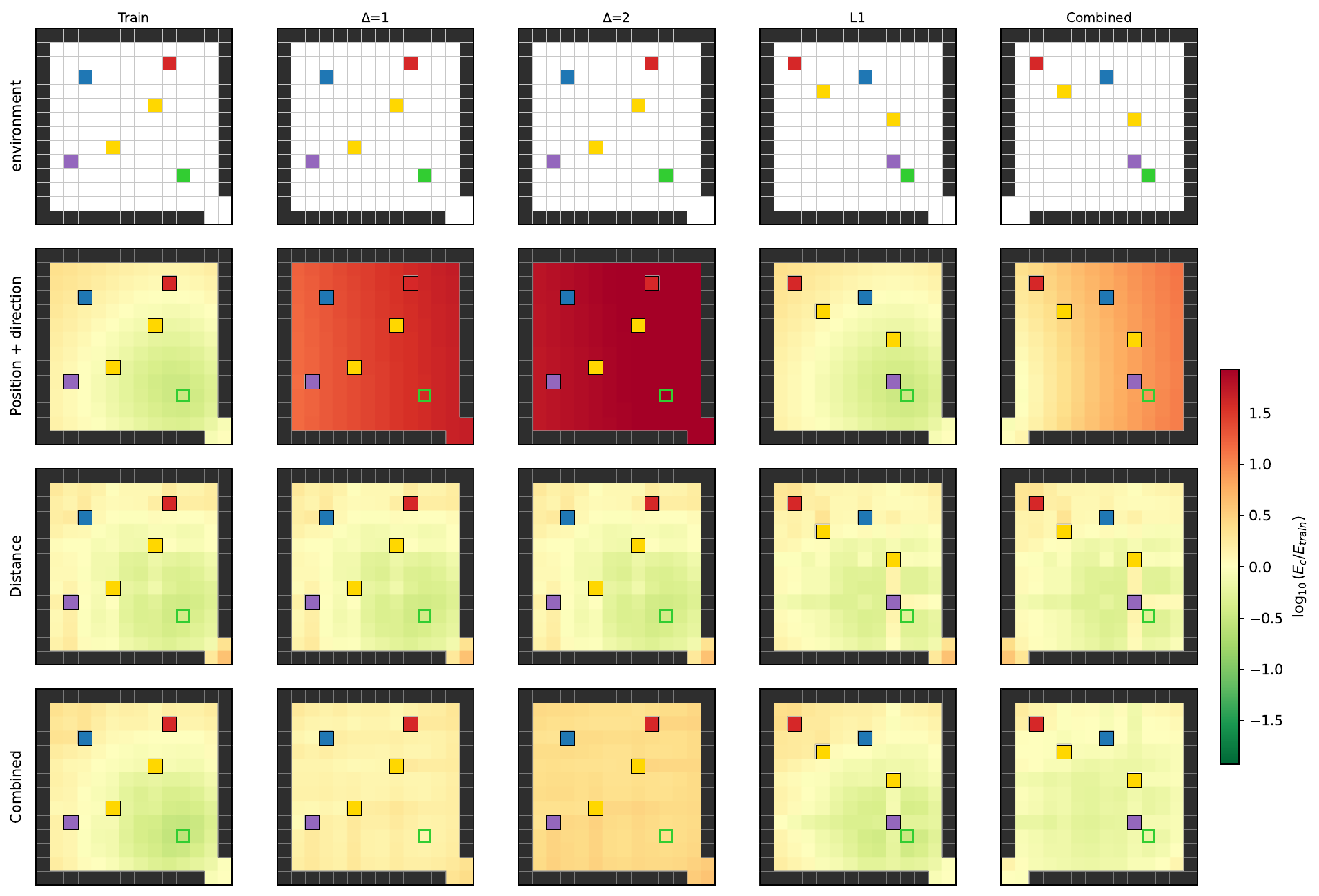}
    \caption{
    Controlled physical-context shifts for TD-MPC across the three observation
    representations. The experiments separate translation, layout change, and
    their combination while keeping the underlying action and reward rules
    fixed. Colors show the one-step dynamics generalization gap.
    }
    \label{fig:appendix_tdmpc_contexts}
\end{figure}

The FourRooms results in the main text show that generalization depends strongly
on the observation representation, but the unseen rooms change both spatial
location and layout. We therefore use controlled physical-context shifts to
separate these effects. Figures~\ref{fig:appendix_dyna_contexts} and
\ref{fig:appendix_tdmpc_contexts} show the results for Dyna-Q and TD-MPC,
respectively.

The translation conditions provide the cleanest comparison because the layout,
obstacles, action rules, and reward mechanism are unchanged. Only the global
spatial realization differs from training. Position + direction observations
become increasingly sensitive to these shifts, while the distance-based
representation remains considerably more stable. This supports the main-text
observation that generalization depends on whether the representation
contains information tied to a particular physical context.

The layout condition tests a different source of change. Here the global
spatial region remains the same, but obstacle locations differ from training.
Errors around changed obstacles are expected because their new locations are
not explicitly represented by position + direction observations. This case
therefore measures both reuse of the shared motion rules and the difficulty of
predicting transitions under previously unseen geometry.

The combined (layout + translation) condition combines both sources of difficulty. The
model must operate in a new spatial region while also encountering a different
local geometry. We therefore interpret this condition as a harder physical
context change rather than as a pure test of coordinate extrapolation.

\subsection{TD-MPC latent representation diagnostics}
\label{linear}
Because TD-MPC predicts dynamics in a learned latent space, we further examine
whether the learned representation preserves useful information across
contexts. Table~\ref{tab:probe_settings} summarizes the training targets and
evaluation settings for the representation diagnostics. Direction and
local-position probes are trained only on the training context and then
evaluated unchanged in unseen contexts. Context decodability is instead a
post-hoc diagnostic trained across contexts to test whether context identity
can be recovered from the latent representation.

For all probe plots, individual points correspond to independently trained seeds, while the larger marker and error bar show the mean and 95\% confidence interval across seeds. Below, we describe each probing experiment in more details.

\begin{table}[ht]
\caption{Summary of the TD-MPC latent-space diagnostics.}
\centering
\small
\begin{tabular}{llll}
\toprule
Diagnostic & Probe input $X$ & Target $y$ & Metric \\
\midrule
Context decoding
& latent $z$
& room identity
& accuracy \\

Direction transfer
& latent $z$
& agent direction
& accuracy \\

Position transfer
& latent $z$
& normalized local $(x,y)$
& $R^2$ \\

Latent alignment
& matched latent sets $(Z_0,Z_c)$
& -- 
& Linear CKA \\
\bottomrule
\end{tabular}
\label{tab:probe_settings}
\end{table}

\begin{figure}[ht]
    \centering
    \includegraphics[width=\linewidth]
    {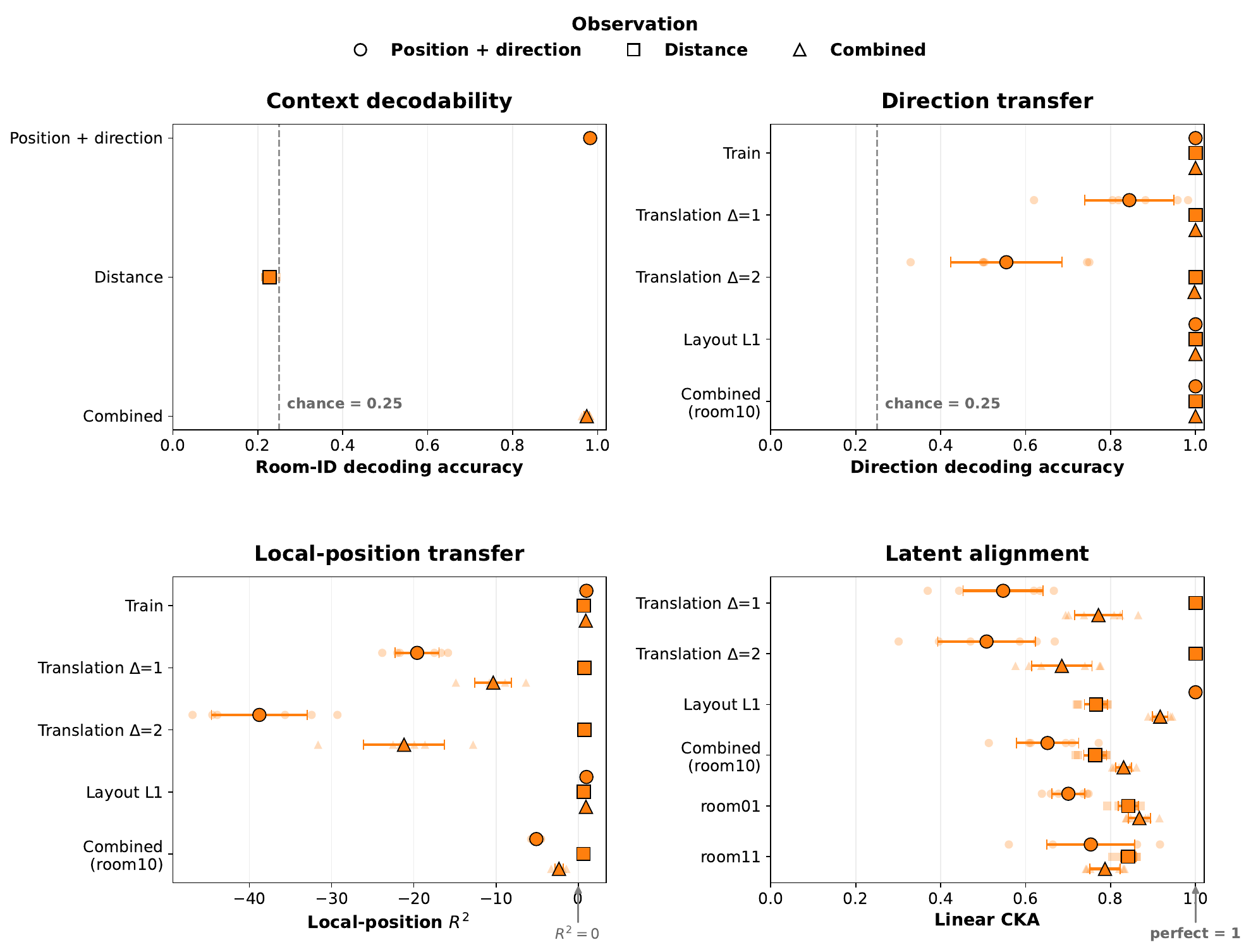}
    \caption{
    \textbf{Diagnostics of TD-MPC's learned latent representations.}
    Direction and position probes
    are trained on the training context and transferred to unseen contexts
    without updating their parameters. Context decoding instead measures whether
    room identity is linearly recoverable from the learned representation.
    Points show individual seeds and larger markers show the mean with 95\%
    confidence intervals. The results show that representation generalization
    depends strongly on the observation mode, with the largest failures under
    spatial translation occurring when the representation is built from
    absolute position information.
    }
    \label{fig:appendix_latent_probes}
\end{figure}

\paragraph{Context decodability.}
We first ask whether the latent representation contains enough information to identify the room in which a state was observed. We pool latent vectors from the four base rooms (\texttt{room00}, \texttt{room10}, \texttt{room01}, and \texttt{room11}) and use the latent vector as input, $X=z$, and the room identity as the classification target. We use an 80/20 stratified train/test split with random seed 42. The probe is a standardization layer followed by logistic regression with a maximum of 3000 iterations. Since there are four rooms, chance accuracy is $0.25$. This probe measures whether context identity is linearly decodable from the latent representation; it is not a cross-context transfer test.

\paragraph{Direction transfer.}
We next ask whether the agent's direction is represented in a way that remains stable across contexts. A linear classifier is trained only on latent vectors from the training context, \texttt{room00}, with $X=z$ and the four-valued agent direction as the target. We again use an 80/20 split, stratified by direction, with random seed 42. The classifier consists of standardization followed by logistic regression with a maximum of 3000 iterations. We first report accuracy on the held-out part of \texttt{room00}. We then refit the probe using all \texttt{room00} states and evaluate it on the unseen contexts without updating the probe. Chance accuracy is $0.25$.

\paragraph{Local-position transfer.}
We similarly test whether local position is encoded consistently across contexts. The probe input is again the latent vector, $X=z$, while the regression target is the normalized local position, $y=(x_{\mathrm{local}}/S,y_{\mathrm{local}}/S)$, where $S$ is the room size. The same 80/20 split used for the direction probe is used here. We fit a standardization layer followed by ridge regression with regularization $\alpha=1$. After measuring performance on the held-out training states, the probe is refitted using all training-room states and transferred unchanged to the unseen contexts. We report the coefficient of determination, $R^2$, in Fig.~\ref{fig:appendix_latent_probes}; $R^2=0$ corresponds to the constant-mean baseline, while negative values mean that the transferred probe performs worse than this baseline.

\paragraph{Latent alignment.}
The final diagnostic does not train a probe. Instead, we directly compare the geometry of latent representations between the training context and each unseen context. We match states that have the same local position and agent direction in both contexts and compare their latent matrices. We use linear centered kernel alignment (CKA), a similarity measure between two representations. A value near $1$ indicates that the matched latent representations preserve very similar geometry, while lower values indicate that the representation changes across contexts. Only states that are valid in both contexts are included in this comparison.

Overall, the diagnostics show that learning a latent representation does not
automatically make it invariant to physical context. The amount and stability
of semantic information in TD-MPC's latent space depend on the observation
representation and the evaluation context. At the same time, the stronger
results with distance-based observations show that reliable reuse is possible
when the input already removes much of the irrelevant physical variation.
These results motivate representations that preserve shared behavioral
structure while reducing sensitivity to physical changes that are not needed
for predicting the reusable dynamics.

\section{Additional related work}
This section reviews broader work in continual reinforcement learning, its
intersection with model-based reinforcement learning, and zero-shot
generalization.

\subsection{Model-based RL}
Model-based reinforcement learning (MBRL) has recently shown strong performance
across a wide range of single-task problems, making it a promising direction
for building more general agents~\citep{m2023model}.
Dyna~\citep{dyna} as a classic algorithm, and Dreamer~\citep{hafner2025dreamerv3}, TD-MPC~\citep{hansen2022tdmpc}, and methods
based on joint-embedding predictive architectures (JEPA)~\citep{leworldmodel2026}
are examples of this growing family. These methods learn an internal model of
the environment and use it for prediction, planning, or imagined experience.
They mainly differ in how the world model is represented and how it is used for
decision making. However, model errors can affect policy improvement and
planning~\citep{janner2021trustmodelmodelbasedpolicy}. Most work in this
direction focuses on strong performance within a single task, while reuse and
generalization across a sequence of changing tasks have received less
attention.

\subsection{Continual model-based RL}
Work at the intersection of continual reinforcement learning and model-based
reinforcement learning has explored several ways of preserving and transferring
knowledge across tasks. \emph{Replay-based approaches} modify how previous experience is stored and sampled from the replay buffer, with the goal of reducing catastrophic forgetting while preserving plasticity for new tasks \citep{rahimi2023replay, pmlr-v232-kessler23a, drago, alyahya2026arrow}. \emph{Dual-architecture approaches} maintain two different models, often with task-specific and task-agnostic roles, and transfer information between them when a new task is encountered \citep{huang2021continual, fu2022model}. More recently, \citet{pmlr-v267-liu25p} study continual RL directly through
online world model learning and model predictive control, with the goal of
learning new tasks while avoiding catastrophic forgetting.

More broadly, continual reinforcement learning has also been studied through \emph{regularization-based methods}, which constrain parameter updates to preserve previously learned knowledge \citep{kirkpatrick2017overcoming}, and \emph{network-expansion methods}, which increase model capacity as new tasks arrive and dedicate additional parameters to learning them \citep{schwarz2018progress}.

Many of these approaches assume that task boundaries are known. In contrast, a learned world model may provide useful signals for detecting changes when new observations become inconsistent with its predictions. Dual-architecture methods are particularly related to our perspective, since a
task-agnostic component may support transfer and retention across tasks.
However, our view also asks whether a new learning phase is needed at all.
When previously learned behavioral knowledge is still applicable, the world
model should ideally be reused directly rather than relearned or adapted
unnecessarily.

\subsection {Zero-shot generalization in (deep) RL}
Prior studies of the zero-shot generalization problem in RL agents mainly focus on model-free approaches~\cite{Kirk_2023}. Most of the work in this direction tries to incorporate a definition of `context' (long-term properties of the environment, which is fixed during a learning episode) in their methods and model their problem in a Contextual Markov Decision Process (CMDP)~\cite{hallak2015contextual}. Agents learn a task under a specific defined context. To measure an agent's generalization, it is then tested in turn under an unseen new context. Train and test contexts in most cases share structure and similar properties, and the agent is supposed to leverage this shared structure from the learned task to unseen ones. Methods in this direction mostly differ in how they try to use the context and also their access level to context in the environment (fully observable or they need to be inferred). Recently, model-based approaches have also considered this setting.
\citet{prasanna2024dreamingworldslearningcontextual} assume access to context
information and, after observing weak zero-shot generalization of Dreamer,
introduce a contextual version of the model. \citet{roder2026dynamics} instead
infer context through a self-supervised module. Related work has also used
context inference with policy-learning methods
~\citep{lee2020contextawaredynamicsmodelgeneralization} or hypernetworks that
condition model parameters on observed context
~\citep{beukman2023dynamicsgeneralisationreinforcementlearning}.

Our focus is complementary to these approaches. Rather than proposing a
context-aware mechanism, we study why previously learned world-model knowledge
may fail to transfer. We separate physical and behavioral context and use
controlled changes in physical context and observation representation to study
which factors limit zero-shot reuse in continual RL.

Another line of work studies invariant representation learning. These methods
aim to preserve task-relevant information while reducing sensitivity to
variations that should not affect the agent's behavior. This can improve
generalization when different tasks share common structure.
\citet{shimizu2024bisimulationmetricmodelpredictive}, for example, introduce
bisimulation-based representation learning for TD-MPC to encourage more robust
latent representations. However, the zero-shot generalization of this method
across contextual changes has not been studied directly. This direction is closely
related to our observation that reliable reuse depends strongly on whether the
representation preserves shared behavioral structure across physical-context
changes.


\end{document}